\documentclass[conference]{IEEEtran}

\usepackage{cite}

\ifCLASSINFOpdf
   \usepackage[pdftex]{graphicx}
\else
\fi
\usepackage{amsmath}
\usepackage{amssymb} 
\usepackage{amsthm,bm}
\usepackage{dsfont}
\usepackage{mathrsfs}

\usepackage{color,colortbl}
\usepackage{float}
\usepackage{subfigure}
\usepackage{caption}
\usepackage{tikz}
\usetikzlibrary{shapes,arrows,positioning,calc,backgrounds}
\usetikzlibrary{shapes.multipart}

\usepackage{siunitx} 

\usepackage{makecell}
\usepackage{multirow}

\newcommand{\vect}[1]{\mathbf{#1}}

\DeclareMathOperator*{\argmax}{arg\,max}
\DeclareMathOperator*{\argmin}{arg\,min}

\newcommand{\nlenv}[1]{\begin{tabular}{c} 
    #1
  \end{tabular}
}

\definecolor{lightgray}{rgb}{0.83, 0.83, 0.83}

\usepackage{enumitem}

\usepackage{todonotes}

\begin{document}

\title{Transmitter Discovery through\\ Radio-Visual Probabilistic Active Sensing}

\author{Luca Varotto and Angelo Cenedese
\thanks{L. Varotto and A. Cenedese are with Department of Information Engineering,
        University of Padova, Italy. Email:
        {\tt luca.varotto.5@phd.unipd.it}}%
}
\author{
\IEEEauthorblockN{Luca Varotto}
\IEEEauthorblockA{Department of Information Engineering\\
University of Padova, Italy\\
Email: luca.varotto.5@phd.unipd.it}
\and
\IEEEauthorblockN{Angelo Cenedese}
\IEEEauthorblockA{Department of Information Engineering\\
University of Padova, Italy\\
Email: angelo.cenedese@unipd.it}
}


%


\maketitle

\begin{abstract}
Multi-modal Probabilistic Active Sensing (MMPAS) uses sensor fusion and probabilistic models to control the perception process of robotic sensing platforms. MMPAS is successfully employed in environmental exploration, collaborative  mobile  robotics, and target tracking, being fostered by the high performance guarantees on autonomous perception.
In this context, we propose a bi-Radio-Visual PAS scheme (Ra$^2$ViPAS) to solve the transmitter discovery problem. Specifically, we firstly exploit the correlation between radio and visual measurements to learn a target detection model in a self-supervised manner. Then, the model is combined with antenna radiation anisotropies into a Bayesian Optimization framework that controls the platform. 
We show that the proposed algorithm attains an accuracy of $92\%$, overcoming two other probabilistic active sensing baselines.
\end{abstract}


%
\IEEEpeerreviewmaketitle

\section{Introduction}
\textit{Active sensing}~(AS)~\cite{radmard2017active}
consists in the control of a dynamical system with actuation and sensing capabilities (e.g., eye-in-hand manipulators~\cite{radmard2017active}), with the purpose of automating the perception process and maximizing its efficiency. 
By leveraging the intimate interplay among estimation, perception, and control, AS stands at the nexus between automation and robotics research, fostered by 
the higher performance guarantees provided by actuated sensing platforms w.r.t passive counterparts~\cite{active_classification}.
Indeed, by enabling autonomous perception in robotics systems, AS is successfully employed in 
environmental exploration~\cite{popovic2019enviromental}
collaborative mobile robotics~\cite{ghassemi2020extended}
target tracking~\cite{varotto2020probabilistic}, 
and source seeking~\cite{park2020cooperative}.

Along this line, \textit{probabilistic}~AS (PAS)~\cite{queralta2020collaborative}
exploits incoming data to generate a belief map. This encodes the knowledge gathered during the sensing mission (e.g., target potential locations in tracking tasks~\cite{radmard2017active}),
and
is used to guide the platform towards the next actions.
More importantly, probabilistic approaches account for realistic perception uncertainties~\cite{meera2019obstacle}; hence, they are suitable to manage real-world (noisy) scenarios, unmodeled dynamics, sensing nuisance.
Furthermore, probabilistic decision making has high adaptivity properties~\cite{popovic2019enviromental}, and
it is useful when poor a-priori knowledge is available~\cite{ramirez2014moving}.

Novel technologies and the evolution in embedded systems have enabled the integration of different sensing modalities in autonomous robotic platforms, 
paving the way towards \textit{multi-modal} PAS (i.e., MMPAS)~\cite{varotto2020probabilistic}.
The coupling of different information sources opens up new perspectives in scene perception: multi-sensor platforms enable 
parallelisation and specialisation~\cite{esterle2017future}, while sensors heterogeneity induces inherent robustness and complementarity (i.e., different properties of the environment can be perceived)~\cite{varotto2020probabilistic}. Notably, the aggregated data allow inferences that are not possible with single-sensor measurements.

In this context, \emph{Radio-Visual PAS} (RaViPAS) approaches have been recently designed to mitigate the inadequacies of camera and radio-only strategies~\cite{varotto2020probabilistic}, especially in localization tasks~\cite{de2017efficient}.
Indeed, the information richness of visual sensors 
may be impaired by
occlusions and Field of View (FoV) directionality~\cite{de2017efficient}.
Radio frequency (RF) signals, instead, have wider ranges~\cite{zanella2016best} and enable energy efficient localization from the extraction of the Received Signal Strength Indicator (RSSI) of standard packet traffic~\cite{zanella2016best}. Furthermore, RF communication 
has low hardware requirements and comes as parasitic in many real-life scenarios, since most portable devices are WiFi or Bluetooth enabled. Nonetheless, environmental interference 
often limits the RF-based localization accuracy~\cite{zanella2016best}.
RaViPAS aims to combine the complementary benefits of RF signals with visual cues.
The literature addressing radio-visual sensor fusion is still sparse and the RaViPAS framework 
is an open research field with methodological challenges and application opportunities: for example, energy-aware strategies are required to alternate the accurate but energy-harvesting camera measurements, with the rough but lighter radio observations~\cite{varotto2021probabilistic,de2017efficient}. Furthermore, traditional \mbox{RF-based} solutions involve tiring human-labeled calibration processes~\cite{zanella2016best}.
In this regard, the integration of accurate camera measurements in self-supervised calibration methods have been proven to be beneficial~\cite{de2017efficient,varotto2021probabilistic}.
%

\noindent \textbf{Contribution -}
In this work, we propose a \mbox{RaViPAS} scheme
to solve the \textit{transmitter discovery} problem hereafter described \mbox{(Fig. \ref{fig:scenario})}, which, despite its apparent simplicity, remains a canonical problem in many application domains. A static radio-visual sensing platform  
is surrounded by targets whose number and location is initially unknown. Only one of them establishes communication with the platform through a radio transmitter (Tx) and the aim is to identify which target is the Tx.
We design a Bayesian Optimization (BO) controller, leveraging the non-isotropic antenna radiation pattern at the receiver side. 
The exploration-exploitation trade off capabilities of BO are crucial for robust and efficient target localization~\cite{meera2019obstacle}.
In addition, 
RSSI-based localization techniques hinging on BO~\cite{carpin2015uavs}, do not require any observation model. Hence, the tiring human-labeled calibration processes involved in traditional RF-based solutions can be avoided.  
To increase the robustness of the localization process, we include a target visual detectability model into the BO framework. To this aim, we devise a self-supervised training procedure, exploiting the correlation between radio-visual inputs. In this way, we automatize the dataset acquisition process and minimize human intervention. In accordance with PAS paradigm, the platform is guided by a probabilistic reward function,
which is refined as new observations are collected. Numerical results show that the proposed algorithm overcomes a RF-only PAS scheme.

\section{Problem Statement}\label{sec:problem_statement}
\begin{figure}[t!]
\centering
\includegraphics[width=0.4\textwidth]{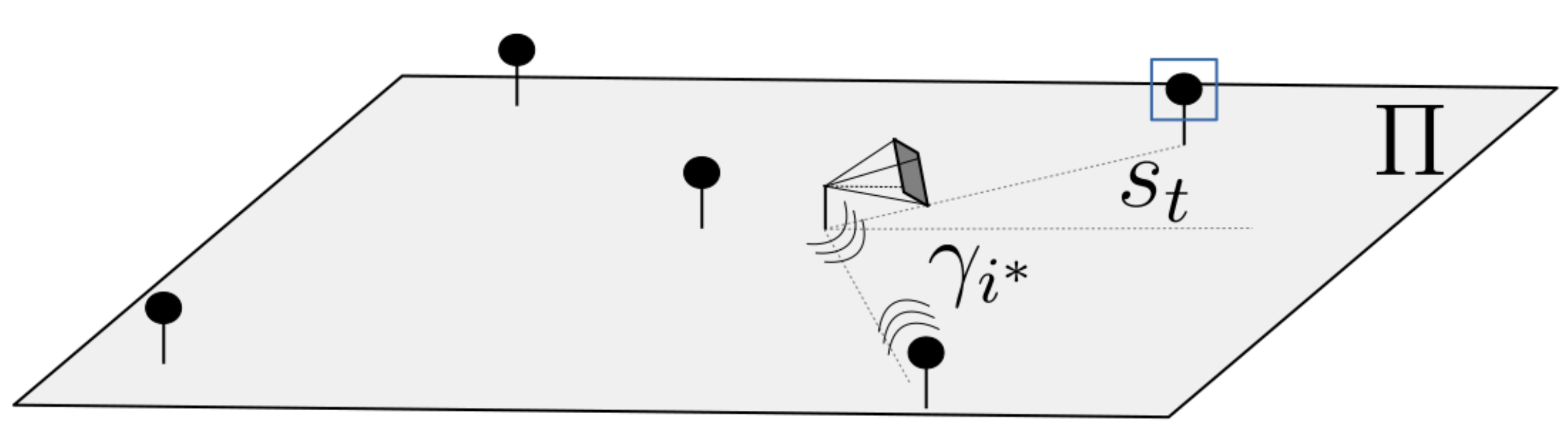}
\caption{\footnotesize Problem scenario. Five targets (black markers) surround the sensing platform; this communicates with one of the targets (Tx) through radio signals. The platform, at orientation $s_t$, detects a target (blue box). The Tx target can be detected when $s_t=\gamma_{i^*}$. 
}
\label{fig:scenario}
\vspace{-0.5cm}
\end{figure}

Fig.~\ref{fig:scenario} shows the main elements of the problem scenario, namely the targets and the sensing platform\footnote{Bold letters indicate (column) vectors if lowercase, matrices otherwise. A Gaussian distribution over the random variable $x$ with expectation $\mu$ and variance $\sigma^2$ is denoted as $\mathcal{N}(x|\mu,\sigma^2)$. 
A Bernoulli distribution with parameter $p$ is denoted as $\mathcal{B}(p)$. 
The shorthand notation $z_{t_0:t_1}$ indicates a sequence 
$\left\lbrace z_k \right\rbrace_{k=t_0}^{t_1}$.}.

\subsubsection*{\textbf{Sensing platform}}

The sensing platform is a static orientable
smart camera~\cite{varotto2020probabilistic}, 
endowed with a real-time target detector~\cite{yolo}
and two radio receivers~\cite{nRF52832_datasheet}.

\emph{The camera:}
Let $\mathcal{F}_0$ be the global reference frame, whose origin is centered in the platform position and whose axis $Z_0$ is orthogonal to the groundplane $\Pi \subset \mathbb{R}^2$ where the platform and the targets lie. We identify the platform state with its only degree of freedom, that is the camera pan angle (i.e., the orientation around $Z_0$)   
\begin{equation}\label{eq:platform_state}
    s_t \in [ -\pi,\pi ], \quad t=LT_c, \; L \in \mathbb{N},
\end{equation}
where $T_c$ is the camera frame rate.  
The camera pan angle defines the Field of View (FoV), $\Phi(s_t) \subset \Pi$, which 
is regulated through the control input $u_t$ according to a deterministic Markovian transition model
\begin{equation}\label{eq:platform_dynamics}
    s_{t+T_c} = s_t + u_t, \quad u_t \in [ -\pi,\pi ].
\end{equation}
The state transitions occur at multiple of $T>T_c$, namely
\begin{equation}\label{eq:input_transition}
    u_t = 0, \quad t \neq HT, \; H \in \mathbb{N}.
\end{equation}

\emph{The receivers:}
The platform is endowed with two radio channels. The former, referred to as Rx$^{(iso)}$, uses an omnidirectional antenna, while the latter, Rx$^{(dir)}$, has a directional antenna~\cite{gomez2018hybrid}.
The dominating lobe of Rx$^{(dir)}$ is supposed to be aligned with the camera optical axis, but the knowledge on the overall radiation pattern of is inaccurate.
We set
\begin{equation}\label{eq:sampling_time_relation}
    T=T_{\text{RF}} = \nu T_c, \; \nu \in \mathbb{N},
\end{equation}
where $T_{\text{RF}}$ is the sampling interval of both receivers. With the first equality ($T=T_{\text{RF}}$), a control input is computed as a new RSSI sample is collected. The second equality ($T_{\text{RF}} = \nu T_c$) is justified by the fact that radio reception has typically longer sample rates than cameras' acquisition~\cite{vollmer2011high}. 

\subsubsection*{\textbf{Targets}} 
The platform is surrounded by $N$ targets, which can be thought as people or marked objects, in civil or industrial settings. The targets are in Line-Of-Sight (LOS) w.r.t. the platform. The position of the $i$-th target w.r.t. $\mathcal{F}_0$ is denoted as $\vect{p}_i \in \Pi$. Accordingly, the distance of the $i$-th target to the platform is the Euclidean norm of $\vect{p}_i$ (i.e., \mbox{$d_i = \lVert \vect{p}_i \rVert_2$}).
The number and the locations of the targets are unknown to the platform, but all $N$ targets can be recognized through the camera object detector; moreover, one of the $N$ targets establishes communication with the platform through an omnidirectional radio Tx. We denote with {$i^* \! \in \! \{ \! 1,\! \dots, \!N \! \}$} the index of such target.

\subsubsection*{\textbf{Perception modeling}} 
Both platform receivers extract from received data packets the Received Signal Strength Indicator (RSSI), $r\in\mathbb{R}$. The RSSI is a measure
of the received radio signal power~\cite{zanella2016best} and it is theoretically related to the Tx-Rx distance $d_{i^*}$~\cite{zanella2016best}, namely 
\begin{equation}\label{eq:d_RSSI}
    r^{(iso)} = g(d_{i^*})
\end{equation}
where $r^{(iso)}$ denotes the RSSI collected at Rx$^{(iso)}$. 
To account for the non-isotropic pattern of the  Rx$^{(dir)}$, the relation \eqref{eq:d_RSSI} should be modified as
\begin{equation}\label{eq:radiation_pattern}
    r_t^{(dir)} = g(d_{i^*})\varrho(s_t,\gamma_{i^*}), \; \varrho(s_t,\gamma_{i^*}) \in [0,1],
\end{equation}
where $\gamma_{i^*}$ is the angular position of Tx w.r.t $\mathcal{F}_0$, that is 
\begin{equation}\label{eq:gamma}
    \gamma_{i^*} = \arctan \left( \frac{p_{i^*,Y_0}}{p_{i^*,X_0}} \right)
\end{equation}
with $p_{i^*,X_0}$ and $p_{i^*,Y_0}$ $X_0$ and $Y_0$ coordinates of $\vect{p}_{i^*}$, respectively. The function $\varrho(\cdot)$ is an attenuation factor that accounts for radiation pattern anisotropies through the misalignment of Tx and $\text{Rx}^{(dir)}$. Indeed, the attenuation is minimal (i.e., $\varrho(s_t,\gamma_{i^*}) \approx 1$) if $s_t \approx \gamma_{i^*}$, that is, if Tx is along the axis of the dominating lobe. The attenuation increases together with the Tx-$\text{Rx}^{(dir)}$ misalignment~\cite{gomez2018hybrid}. 

Motivated by the energy efficiency and the low-cost hardware requirements, RSSI-based localization systems have been widely studied in literature~\cite{de2017efficient,varotto2020probabilistic}. Despite this, RSSI localization suffers from several drawbacks. 
At first, the functions $g(\cdot)$ and $\varrho(\cdot)$ are usually unknown and need to be estimated through extensive and time-consuming calibration procedures~\cite{zanella2016best}. 
In this work, we propose a self-supervised methodology that requires minimal human intervention and does not need to estimate neither $g(\cdot)$, nor $\varrho(\cdot)$ (see \mbox{Sec. \ref{sec:method}}).
A further limitation of RF-only localization systems resides on the high
sensitivity to environmental interference (e.g., cluttering and multi-path distortions)~\cite{zanella2016best}.  
Hence, the target position is hidden in extremely noisy receiver measurements, according to the following {\em RSSI observation model}
\begin{equation}\label{eq:obs_Rx}
\begin{split}
    & z_{\text{RF},t} =
r(\vect{p}_{i^*}) + v_{\text{RF},t} , \quad t = M T_{\text{RF}}, \; M \in \mathbb{N} \\
& v_{\text{RF},t} \sim \mathcal{N}\left(v| 0,\sigma_{\text{RF}}^2 \right)
\end{split}
\end{equation} 
where $z_{\text{RF},t}$ is denoted either as $z_{\text{RF},t}^{(iso)}$ or $z_{\text{RF},t}^{(dir)}$, if $r(\vect{p}_{i^*})$ follows \eqref{eq:d_RSSI} or \eqref{eq:radiation_pattern}.
%
A final issue of RSSI-based localization regards those techniques relying on omnidirectional receivers only~\cite{shahidian2017single}. As \eqref{eq:d_RSSI} highlights, the RSSI values are characterized by inherent ambiguities on the target position (i.e., they do not identify uniquely the position of the radio source).
When the receiver is static, these ambiguities induce severe convergence issues in the  estimation process~\cite{varotto2020probabilistic}.
For this reason, most literature solutions mitigate the ambiguity effects through multiple receivers~\cite{canton2017bluetooth}, multi-modal (e.g., radio-visual) perception systems~\cite{varotto2020probabilistic}, or active sensing schemes~\cite{shahidian2017single}.

As for the camera sensing capability, the visual detection process of target $i$ is modeled as a Bernoulli random variable $D_t \in \{0,1\}$, 
with success probability~\cite{varotto2020probabilistic}
\begin{equation}\label{eq:POD}
    p(D_t=1|\vect{p}_i,s_t) =
    \begin{cases}
    p_D(d_i), & \text{if } \vect{p}_i \in \Phi(s_t) \\
    0, & \text{otherwise}.
    \end{cases}
\end{equation}
As \eqref{eq:POD} highlights, the detection success probability is a function of both the camera orientation and the target-platform distance. In particular, visual depth effects~\cite{radmard2017active} are accounted through $p_D(d_i)$, which is the {\em Probability of Detection} (POD) when the target is in FoV. In this work, we suppose that the same POD function can be applied to all targets, which is a reasonable assumption in most practical scenarios.
If $i=i^*$, the dependence on $d_i$ in \eqref{eq:POD} can be equivalently substituted by $r_t$, on the basis of \eqref{eq:d_RSSI}.

\subsubsection*{\textbf{Problem statement}} 
The Tx discovery problem aims at identifying the Tx among the $N$ targets. 
The LOS condition implies
\begin{equation}\label{eq:LOS_condition}
    \nexists (i,j) \text{ s.t. } \gamma_i=\gamma_j,
\end{equation}
that is, each target is uniquely identified by its angular position w.r.t. the platform. Hence, the Tx discovery problem boils down to an \textit{association} task on top of a uni-dimensional \textit{localization} problem within a \textit{noisy} scenario. The quantity to be estimated is the Tx angular position $\gamma_{i^*}$, assuming that the dynamics of the targets is slow w.r.t. the localization process. Formally, at time $t$, the result of the association task is  
\begin{equation}\label{eq:Tx_estimate}
    \hat{i}_t = \argmin_{i \in \{1,\dots,N\}}{ \lvert \hat{\gamma}_t - \gamma_{i} \rvert},
\end{equation}
where $\hat{\gamma}_t$ is the estimate of $\gamma_{i^*}$ (i.e., the localization outcome).

\section{Methodology}\label{sec:method}
Indeed, the Tx discovery problem is ill-posed if tackled with passive strategies, due to the inherent RSSI ambiguities and the uncertainties on the receivers radiation pattern, 
and thus motivated, we propose an active sensing scheme. 

Theoretically, the anisotropy of Rx$^{(dir)}$, coupled with the platform movements, provide sufficient information to reduce the RSSI ambiguities and solve the localization problem. Nonetheless, in practical scenarios the noise in RSSI data affect the stability and reliability of RF-only strategies. Moreover, the lack of knowledge on the antenna radiation pattern often demands extensive calibration procedures. Therefore, we increase the localization robustness by combining radio and visual cues. 
Furthermore, we formulate the problem 
in a Bayesian probabilistic framework, which accounts for perception uncertainties and increases the adaptivity properties.
Overall, we obtain the bi-Radio-Visual Probabilistic Active Sensing (Ra$^2$ViPAS) scheme depicted in Fig.~\ref{fig:pipeline} and detailed in the following. The Ra$^2$ViPAS is based on two steps:
\begin{enumerate}
\item first, we apply Gaussian Process Regression (GPR) to learn the function $p_D(r)$ in a self-supervised fashion (\mbox{Sec. \ref{subsec:GPR}}); 
    \item then, we cast the localization task to a black-box optimization problem for which Bayesian Optimization (BO) is employed (\mbox{Sec. \ref{subsec:Tx_discovery}}). 
\end{enumerate}

\tikzset{
block/.style = {draw, fill=white, rectangle, minimum height=3em, minimum width=3em},
block_transp/.style = {rectangle, minimum height=3em, minimum width=3em},
sum/.style= {draw, fill=white, circle, node distance=1cm},
block_split/.style = {draw, fill=white,minimum height=3em, minimum width=3em,rectangle split, rectangle split horizontal, rectangle split parts=3}}
\begin{figure}[t!]
\center
\begin{tikzpicture}[auto, node distance=3cm,>=latex',scale=0.7, transform shape]
\node [block] (g) {\parbox[h]{7mm}{\rotatebox[origin=c]{90}{\nlenv{environment \\ and target}}}};
\node [block_split, right of =g] (sensing) { 
\nodepart{one}
Rx$^{(dir)}$
\nodepart{two}
Rx$^{(iso)}$
\nodepart{three}
cam};
\node [block, right of = sensing,xshift=3cm] (sensor dynamics) { $s_t = s_{t-T_c} + u_{t-T_c}^*$};
\node [block, below of = sensor dynamics, yshift=1.2cm,xshift=1cm] (time_update_sensor) { $z^{-1}$};
\node [block, below of = sensor dynamics,yshift=-1.75cm,xshift=0.5cm] (control) { $\argmax{ a(s|\mathcal{D}_t) }$};
\node [block, below of = sensing,yshift=0.25cm,xshift=0.95cm] (POD) {$\hat{p}_D(r) \sim \mathcal{GP}$};
\node [sum, below of = POD,yshift=-1cm] (prod) {$\times$};
\node [block, right of = prod, xshift=-0.5cm, minimum width=5em] (GPR) { \nlenv{Probabilistic \\ map}};

\draw [->] (g.east)  --  ($(sensing)+(-2,0)$);
\draw [->] (sensor dynamics.west) -- node[]{$s_t$} ($(sensing)+(2,0)$);
\draw [->] ($(sensor dynamics.west)+(-1,0)$) |- (time_update_sensor.west) ;
\draw [->] (time_update_sensor.east) -- ++ (0.3,0) -- node[]{ $s_{t-T_c},u_{t-T_c}^*$} ++ (0,1.8) -- ($(sensor dynamics.east)$);
\draw [->] ($(time_update_sensor.west) + (-0.5,0)$) -- ($(control.north)+ (-0.47,0)$) ;
\draw [->] (sensing.one south) |-node[xshift=-0.5cm]{ $z_{\text{RF},t}^{(dir)}$} 
($(prod.north)+(-0.3,-0.3)$);
\draw [->] (sensing.two south) --node[xshift=-1cm]{ $z_{\text{RF},t}^{(iso)}$} ($(POD.north)+(-0.7,0)$);
\draw [->] (sensing.three south) --node[]{ $\tilde{p}_{D,t}$} ($(POD.north)+(0.35,0)$);
\draw [->] (POD.south) --node[xshift=-1cm]{$y_{D,t}$}  (prod.north);
\draw [->] (prod.east) --node[xshift=-0.15cm]{$y_t$}  (GPR.west);
\draw [->] (GPR.east) --  (control.west);
\draw [->] ($(control.north)+(0.53,0)$) -- node[]{ $u_t^*$}(time_update_sensor.south);

\begin{scope}[on background layer]
\draw [fill=lightgray,dashed] ($(sensing) + (-2,0.7)$) rectangle ($(sensing) + (2,-0.7)$);
\draw [fill=lightgray,dashed] ($(sensor dynamics.west) + (-1.8,0.7)$) rectangle ($(time_update_sensor.east) + (0.5,-0.7)$);
\draw [fill=lightgray,dashed] ($(GPR) + (-1.5,1)$) rectangle ($(control) + (1.5,-1.5)$);
\draw [dashed] ($(sensing) + (-2.2,1.4)$) rectangle ($(control) + (2,-2)$);
\end{scope}

\node [block_transp, above of = sensor dynamics,yshift=-2cm] (sensor_annotation) { \textit{Platform state}};
\node [block_transp, below of = control, yshift=2.2cm] (control_annotation) { \textit{Control}};
\node [block_transp, above of = sensing,yshift=-2cm] (sensing_annotation) { \textit{Sensing units}};
\node [block_transp, below of = control, yshift=1.25cm] (BO_annotation) { \textit{Bayesian Optimization}};
\node [block_transp, above of = sensor dynamics, yshift=-1.3cm] (sensor_annotation) { \textit{Sensing platform}};
\end{tikzpicture}
\vspace{-0.2cm}
\caption{ 
Ra$^2$ViPAS scheme for the Tx discovery problem.}
\label{fig:pipeline}
\vspace{-0.5cm}
\end{figure}
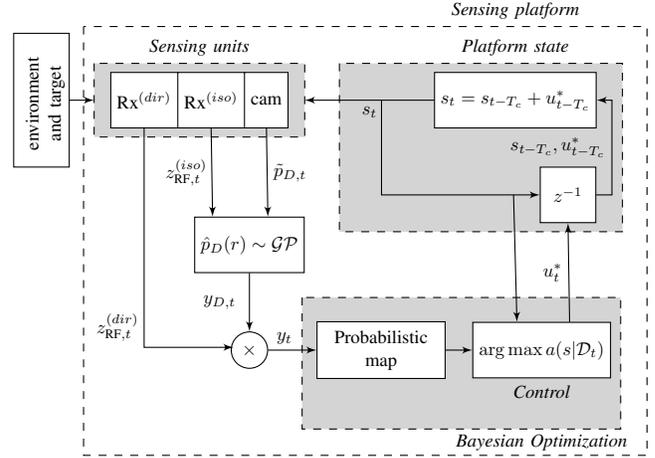
%
%

\subsection{POD learning through GPR}\label{subsec:GPR}

\subsubsection*{\textbf{GPR}}

A Gaussian Process (GP) is a collection of random variables, any finite number of which have a joint Gaussian distribution~\cite{williams2006gaussian}. Given the input vector $\vect{x} \in \mathbb{R}^p$, a GP is fully specified by its mean function $m(\vect{x})$ and covariance function $k(\vect{x},\vect{x}^\prime)$, namely
\begin{equation}\label{eq:GP_prior}
    f(\vect{x}) \sim \mathcal{GP}\left(m(\vect{x}),k(\vect{x},\vect{x}^\prime)\right),
\end{equation}
where $f:\ \mathbb{R}^p \rightarrow \mathbb{R}$ is referred to as {\em latent function}. 
Mean and covariance (or kernel) functions incorporate prior knowledge (e.g., periodicity, smoothness) about the latent function.
The mean function is typically constant (usually zero), while the most commonly-used kernel functions are constant, linear, square exponential or Matern, as well as compositions of multiple kernels~\cite{williams2006gaussian}.
The hyperparameters in the mean and covariance functions are computed by fitting the train dataset $\mathcal{D}$ of cardinality $n_{train}$ 
\begin{equation}\label{eq:dataset}
    \begin{split}
& \mathcal{D} = \left\lbrace (\vect{x}_j,y_j)  \right\rbrace_{j=1}^{n_{train}} = (\vect{X},\vect{y}) \\
    & \vect{X} = 
\begin{bmatrix}
\vect{x}_1 & \dots & \vect{x}_{n_{train}}
\end{bmatrix}^\top \in \mathbb{R}^{n_{train} \times p} \\
    & \vect{y} =
\begin{bmatrix}
y_1 & \dots & y_{n_{train}}
\end{bmatrix}^\top
\in \mathbb{R}^{n_{train}}.
    \end{split}
\end{equation}
Each training label $y_j$ is a noisy measurement of the latent function $f(\vect{x})$, namely
\begin{equation}\label{eq:GP_noisy_labels}
    y_j = f(\vect{x}_j) + \epsilon_j, \quad \epsilon_j \sim \mathcal{N}(\epsilon|0,\sigma^2).
\end{equation}
To account for the i.i.d. Gaussian noise in the training labels, the GP in \eqref{eq:GP_prior} becomes~\cite{williams2006gaussian},
\begin{equation}\label{eq:GP_prior_labelsNoise}
    f(\vect{x}) \sim \mathcal{GP}\left(m(\vect{x}),k(\vect{x},\vect{x}^\prime) + \sigma^2\vect{I}\right).
\end{equation}
This GP is used as \textit{prior} for non-parametric Bayesian inference of the latent function. 
Consider the test inputs 
$
    \vect{X}_* = 
\begin{bmatrix}
\vect{x}_{1,*} & \dots & \vect{x}_{n_{test},*}
\end{bmatrix}^\top \! \in \! \mathbb{R}^{n_{test} \times p}.
$
From the definition of a GP, any finite number of samples drawn from the GP are jointly Gaussian. Thus, according to~\cite{williams2006gaussian},
\begin{equation}\label{eq:GP_joint}
    \begin{bmatrix}
    \vect{y} \\ \vect{y}_*
    \end{bmatrix} \!= \!
    \mathcal{N} \left( 
    \begin{bmatrix}
    \bm{\mu} \\
    \bm{\mu}_*
    \end{bmatrix} \! , \!
    \begin{bmatrix}
    \vect{K}(\vect{X},\vect{X})+\sigma^2\vect{I}  
    \! & \! \vect{K}(\vect{X},\vect{X}_*) \\
    \vect{K}(\vect{X}_*,\vect{X}) \! & \!
    \vect{K}(\vect{X}_*,\vect{X}_*)
    \end{bmatrix}
    \right)
\end{equation}
where: $\vect{y}_*=
\begin{bmatrix}
f(\vect{x}_{1,*}) & \dots & f(\vect{x}_{n_{test},*})
\end{bmatrix}^\top \! \in \! \mathbb{R}^{n_{test}}
$;
the \mbox{$j$-th} row of $\bm{\mu}$ and $\bm{\mu}_*$ is $m(\vect{x}_j)$ and $m(\vect{x}_{j,*})$, respectively; 
the \mbox{$(i,k)$-th} entry of $\vect{K}(\vect{X},\vect{X})$ and $\vect{K}(\vect{X},\vect{X}_*)$ is $k(\vect{x}_j,\vect{x}_k)$ and $k(\vect{x}_j,\vect{x}_{k,*})$, respectively.

Making predictions about unobserved values 
$\vect{X}_*$
consists in drawing samples from the {\em predictive posterior distribution} of $\vect{y}_*$, given $\mathcal{D}$ and $\vect{X}_*$, that is
\begin{equation}\label{eq:GP_posterior}
\begin{split}
    & \vect{y}_*|\mathcal{D},\vect{X}_* \sim \mathcal{N}(\vect{y}_*|\bm{\mu}_{*|\mathcal{D}},\bm{\Sigma}_{*|\mathcal{D}})  \\
    & \bm{\mu}_{*|\mathcal{D}} \! = \! \bm{\mu}_*  \! + \! \vect{K}(\vect{X}_*,\vect{X})\left[ \vect{K}(\vect{X},\vect{X}) \! + \! \sigma^2\vect{I} \right]^{-1}(\vect{y}-\bm{\mu}) \\
    & 
    \begin{split}
        \bm{\Sigma}_{*|\mathcal{D}} 
        & = \vect{K}(\vect{X}_*,\vect{X}_*)\\
        & - \vect{K}(\vect{X}_*,\vect{X})\left[ \vect{K}(\vect{X},\vect{X}) +\sigma^2\vect{I} \right] \vect{K}(\vect{X},\vect{X}_*) 
    \end{split}
\end{split}
\end{equation}


\begin{figure}[t]
\vspace{0.2cm}
\centering
\includegraphics[width=0.4\textwidth]{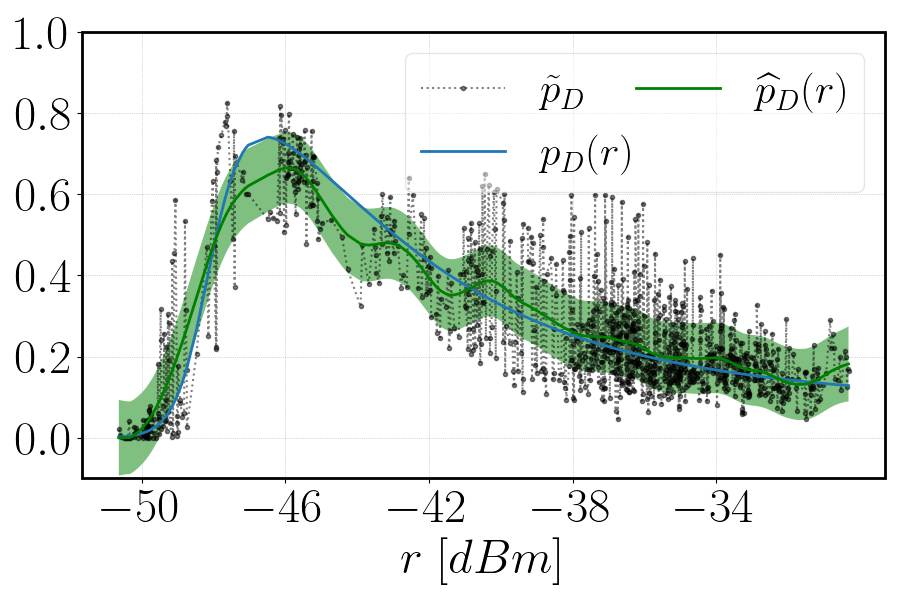}
\caption{Comparison between the latent function $p_D(r)$ (blue line) and the GP model $\widehat{p}_D(r)$ (green), learnt on the very noisy synthetic data $\{z_{\text{RF},i}^{(iso)},\widetilde{p}_{D,i}\}_{i=1}^{n_{train}}$ (black markers).}
\label{fig:pod}
\vspace{-0.2cm}
\end{figure}

\subsubsection*{\textbf{POD learning through GPR}}\label{subsubsec:POD_GPR}
On the basis of \eqref{eq:d_RSSI}, we consider the underlying POD as a function of the RSSI values and we model it as a GP, namely\footnote{The notation $\widehat{p}_D(r)$ is used to distinguish 
the GP model with the underlying POD, $p_D(r)$.} 
\begin{equation}\label{eq:POD_GP}
    \widehat{p}_D(r) \sim \mathcal{GP}\left(0,k(r,r^\prime)\right).
\end{equation}
The train dataset is 
\begin{equation}\label{eq:dataset}
\mathcal{D} = \left\lbrace \left(z_{\text{RF},j}^{(iso)},\widetilde{p}_{D,j} \right) \right\rbrace_{j=1}^{n_{train}}
\end{equation}
where the inputs are RSSI observations from Rx$^{(iso)}$ and the labels are the {\em empirical POD} between two RSSI samples, according to \eqref{eq:sampling_time_relation}, that is
\begin{equation}\label{eq:measured_pD}
\widetilde{p}_{D,j} =  \frac{1}{\nu} \sum_{\ell=1}^{\nu} D_{j,\ell}.
\end{equation}
where $\{D_{j,\ell}\}_{\ell=1}^{\nu}$ are the $\nu$ detection outcomes before $z_{\text{RF},j}^{(iso)}$ is collected.
Notably, $\mathcal{D}$ is automatically acquired by the platform (i.e., no human labeling is required\footnote{To this aim, the camera and the receiver must be synchronized.}).
If $T_{\text{RF}}$ is sufficiently small w.r.t. the target movements, $\{D_{j,\ell}\}_{\ell=1}^{\nu}$ is a sequence of i.i.d. random variables with distribution $\mathcal{B}\left(p_D(r_j)\right)$,
according to \eqref{eq:d_RSSI} and \eqref{eq:POD}.
Then, for a high frame rate camera~\cite{vollmer2011high}, from the Central Limit Theorem, $\widetilde{p}_{D,j}$ converges in distribution to
\begin{equation}\label{eq:d_convergence}
    \mathcal{N}\left(\widetilde{p}_{D,i} \left \vert{ p_D(r_j), \frac{p_D(r_j)(1-p_D(r_j))}{\nu} }\right. \right), \text{ as } \nu \xrightarrow{} \infty.
\end{equation}
Thus, the latent function is related to the noisy labels 
through the generative model
\begin{equation}\label{eq:noisy_labels_pD}
\begin{split}
    & \widetilde{p}_{D,j} = p_D(r_j) + \epsilon_j\\
    & \epsilon_j \sim \mathcal{N}\left(\epsilon \left \vert{0, \frac{p_D(r_j)(1-p_D(r_j))}{\nu}}\right. \right),
\end{split}
\end{equation}
and the GPR problem is well-posed\footnote{
As \eqref{eq:noisy_labels_pD} highlights, the generative model of the labels is input-dependent (i.e., the GP is heteroscedastic~\cite{mchutchon2011gaussian}).
Moreover, the GPR works on noisy inputs, 
according to \eqref{eq:obs_Rx}. To deal with heteroscedasticity and noisy inputs, several ad-hoc methods have been proposed
in literature~\cite{mchutchon2011gaussian}.
}.

Fig. \ref{fig:pod} shows an example of POD learning through GPR over synthetic data (see Sec. \ref{sec:numerical} for the data generation process). The true POD function is
\begin{equation}\label{eq:synthetic_pod}
    p_D(d) = \left[ \left( 1 + e^{4(d-4.5)} \right) \left( 1 + e^{-(d-2.5)} \right) \right]^{-1}.
\end{equation}
The GP model uses a zero mean function and a Matern covariance~\cite{williams2006gaussian}.
To evaluate the regression performance we use the coefficient of determination $R^2 \in (-\infty,1]$
\begin{equation}\label{eq:R_score}
    R^2 = 1 - \frac{\sum_{j=1}^{n_{train}} \left(p_D(r_j) - \widehat{p}_{D,*}(r_j) \right)^2}{\sum_{j=1}^{n_{train}} \left( p_D(r_j) - \frac{1}{n_{train}}\sum_{k=1}^{n_{train}} p_D(r_k) \right)^2}  
\end{equation}
where $\widehat{p}_{D,*}(r_j)$ is the POD prediction, according to \eqref{eq:GP_posterior}.
The higher $R^2$, the better the fit; in the case of Fig. \ref{fig:pod}, $R^2=0.94$.

\subsection{Ra$^2$ViPAS for Transmitter Discovery}\label{subsec:Tx_discovery}

The already introduced Fig. \ref{fig:pipeline} shows the Ra$^2$ViPAS pipeline. The information gathered by the platform sensing channels is aggregated
and injected into a BO scheme. This exploits a GPR sub-module to reconstruct a probabilistic approximation of the localization objective function. This is then used to generate platform control inputs.

\subsubsection*{\textbf{BO}}\label{subsubsec:BO}

BO 
is a procedure designed for 
derivative-free global optimization, particularly suited for objective functions that are expensive to evaluate~\cite{frazier2018tutorial}. Formally, BO aims to solve the following optimization problem
\begin{equation}\label{eq:optimization}
    \vect{x}^* = \argmax_{\vect{x} \in \mathcal{X}} J(\vect{x})
\end{equation}
where $\mathcal{X}$ is a domain space of interest; the objective function $J:\mathcal{X} \rightarrow \mathbb{R}$ is unknown (i.e., black-box optimization), but can be evaluated at any arbitrary query point $\vect{x}_* \in \mathcal{X}$. This evaluation produces a noise-corrupted (stochastic) output \mbox{$y \in \mathbb{R}$}. 
To solve \eqref{eq:optimization}, BO adopts a sequential procedure. At first, the objective function is approximated by a probabilistic model (e.g., a GP), 
easier to optimize and referred as surrogate function. The surrogate model 
is sequentially refined via Bayesian posterior updating (e.g., GPR), as new data 
are collected. To this aim, given the collected dataset $\mathcal{D}_t= \{(\vect{x}_j,y_j)\}_{j=1}^t$, the next query point is chosen according to a suitable selection criterion
$a(\vect{x}|\mathcal{D}_t)$ (acquisition function)
\vspace{-0.2cm}
\begin{equation}\label{eq:acquisition}
    \vect{x}_{t+1} = \argmax_{\vect{x} \in \mathcal{X}} a(\vect{x}|\mathcal{D}_t).
\end{equation}
The acquisition function is designed over the surrogate function to balance exploration with exploitation, and to quantify the utility of a query point to produce a
more informative posterior distribution.
In this work, we have used the GP Upper Confidence Bound (UCB), namely
\begin{equation}\label{eq:ucb}
a(\vect{x}|\mathcal{D}_t) = \mu_{*,\mathcal{D}_t} + \sqrt{\beta} \sigma_{*,\mathcal{D}_t}   
\end{equation}
where \eqref{eq:GP_posterior} is applied, $\sigma_{*,\mathcal{D}_t}$ is $\bm{\Sigma}_{*,\mathcal{D}_t}$ with $n_{test}=1$, and $\beta$ is an exploration-exploitation tuning hyperparameter.



\subsubsection*{\textbf{Probabilistic Controller}}\label{subsubsec:map_controller}

%
The Tx location satisfies 
\begin{equation}\label{eq:optimization_RaViPAS}
\begin{split}
    & \gamma_{i^*} = \argmax_{\gamma \in [-\pi,\pi]} \underbrace{J_D(\gamma)J_{\text{RF}}(\gamma)}_{J(\gamma)} \\
    & J_D(\gamma) = - \lvert p(D|\vect{p}_{i^*},\gamma_{i^*}) - p(D|\vect{p}_{i^*},\gamma) \rvert \\
    & J_{\text{RF}}(\gamma) = \varrho(\gamma,\gamma_{i^*}).
\end{split}
\end{equation}
Specifically: the detectability term, $J_D(s)$, favors those values of $\gamma$ where the target has the same (measured) POD of the Tx; on the other side, the RF term, $J_{\text{RF}}(s)$, accounts for the radiation pattern and gives higher rewards when Rx$^{(dir)}$ is aligned with Tx.
The lack of precise knowledge on $\varrho(\cdot)$ makes \eqref{eq:optimization_RaViPAS} a black-box optimization problem; hence, it can be opportunely solved via BO. At time $t$, the dataset is 
\begin{equation}\label{eq:dataset_BO}
\mathcal{D}_t = \left\lbrace \left( s_j,y_j \right)  \right\rbrace_{j=T_{\text{RF}}}^t 
\end{equation}
with $y_j = y_{D,j}y_{\text{RF},j}$ and
\vspace{-0.2cm}
\begin{equation}
\begin{split}
    & y_{D,j} =  -\lvert \hat{p}_{D,*}(z_{\text{RF},j}^{(iso)}) -  \tilde{p}_{D,j} \rvert   \\
    & y_{\text{RF},j} = \lvert z_{\text{RF},j}^{(dir)} \rvert/\zeta,
\end{split}
\end{equation}
where $\zeta$ is a user-defined scaling hyper-parameter.
Note that, $z_{\text{RF},t}^{(dir)}/\zeta$ is a (noisy) scaled version of $J_{\text{RF}}(\cdot)$,
according to \eqref{eq:obs_Rx} and \eqref{eq:radiation_pattern}; however, this does not affect the optimization process, since the scale is $g(d_{i^*})$, which does not depend on $s_t$. 
Regarding $y_{D,t}$, it is a noisy version of $J_D(\cdot)$, where the noise comes from both the estimation error in $\hat{p}_{D,*}(z_{\text{RF},t}^{(iso)})$ and the measurements error in $\tilde{p}_{D,t}$ (see Sec. \ref{subsubsec:POD_GPR}).
We use the dataset \eqref{eq:dataset_BO}, to approximate $J(\gamma)$ via GPR (see \mbox{Fig. \ref{fig:pipeline}}). The GP model that approximates $J(\gamma)$ is the probabilistic map from which the next control input is computed 
\begin{equation}\label{eq:control_input}
u_t =
\begin{cases}
\argmax_{s\in[-\pi,\pi]} a(s|\mathcal{D}_t) - s_t, & t\!=\!HT_{\text{RF}} \\
0, & \text{otherwise},
\end{cases}
\end{equation}
where condition \eqref{eq:input_transition}, with $T=T_{\text{RF}}$, is taken into account. Finally, the estimate of $\gamma_{i^*}$ at time $t$ is the maximum of the GP predictive posterior mean \eqref{eq:GP_posterior}, namely
\begin{equation}\label{eq:gamma_hat}
    \hat{\gamma}_t = \argmax_{\gamma \in [-\pi,\pi]} \mu_{*|\mathcal{D}_t}
\end{equation}

\section{Numerical Results}\label{sec:numerical}

To evaluate the proposed approach, we consider a Python-based synthetic environment\footnote{https://github.com/luca-varotto/Tx-discovery}.
Sec.~\ref{subsec:params} describes the main setup parameters, as well as the synthetic data generation process. Sec.~\ref{subsec:assessment} defines the metrics used for performance assessment and the baselines considered for comparison. 
The numerical simulation results are discussed in  Sec.~\ref{subsec:discussion_numerical}.

\subsection{Setup parameters}\label{subsec:params}

\begin{table}[t!]
\vspace{0.3cm}
\centering
\caption{Setup parameters for the MC experiment.}
\label{tab:setup_parameters}
\begin{tabular}{c| c}
\hline
\bf{Parameter} & \bf{Value} \\
\hline
\rowcolor{lightgray}
$T_{\text{RF}}$ & $\SI{0.1}{\s}$\\
$\nu$ & $10$\\
\rowcolor{lightgray}
$n_{train}$ & $900$\\
$n_{test}$ & $120$\\
\rowcolor{lightgray}
$N_{tests}$ & $50$\\
$N$ & $20$\\
\rowcolor{lightgray}
$f(d_t,\eta_t)$ & $d_t+\eta_t \quad$
$\eta_t \sim \mathcal{N}\left(0,0.04\right)$ \\
$g(d_{i^*})$ & \makecell{ $\kappa -10n\log_{10}(d_t/\delta)$ 
$\; \kappa\!=\!\SI{-30}{dBm}, \; n\!=\!2, \; \delta\!=\!\SI{1}{\m}$} \\
\rowcolor{lightgray}
$\varrho(s_t,\gamma_{i^*})$ & $1-0.5(s_t-\gamma_{i^*})^2$\\
$\sigma_{\text{RF}}$ & $\SI{3}{dBm}$ \\
\rowcolor{lightgray}
$p_D(d)$ & $\left[ \left( 1 + e^{4(d-4.5)} \right) \left( 1 + e^{-(d-2.5)} \right) \right]^{-1}$\\
\hline
\end{tabular}
\vspace{-0.3cm}
\end{table}


To carry out realistic simulations, most parameters reflect real device characteristics. 
Tx and Rx$^{(iso)}$ are supposed to be equipped with Nordic nRF52832 SoCs~\cite{nRF52832_datasheet}; Rx$^{(dir)}$ differs from Rx$^{(iso)}$ in that the attenuation gain follows the function in \mbox{Tab. \ref{tab:setup_parameters}}~\cite{gomez2018hybrid}.
The function $g(\cdot)$ follows the
log-distance Path Loss Model (PLM) with noise level \mbox{$\sigma_{\text{RF}}=\SI{3}{dBm}$}~\cite{zanella2016best}.
The radio sampling time is set to $T_{\text{RF}}=\SI{0.1}{\s}$, and the value of $\nu$ is set to $10$, as a reasonable trade-off between typical real-life values and the ideal condition stated in~\eqref{eq:d_convergence}.
The underlying POD function is \eqref{eq:synthetic_pod}, designed according to real-life experiments~\cite{varotto2021probabilistic}. 
%
%
During the POD learning phase, the target is supposed to move according to the stochastic linear model reported in Tab.~\ref{tab:setup_parameters}.
Finally, we test the proposed approach in a crowded environment with $N=20$ targets.

\subsection{Performance assessment}\label{subsec:assessment}

To capture the performance variability, numerical evaluation is performed through a Monte Carlo (MC) experiment, composed by $N_{tests} = 50$ tests of duration $T_W=n_{test}T_{\text{RF}}$ each, where $n_{test}=120$ is the number of RSSI collected. At each MC test the position of the $N$ targets 
is randomly chosen.

\subsubsection*{\textbf{Performance metric}}\label{subsubsec:metrics}
The performance of the Tx discovery task is evaluated through the \textit{Discovery Rate}
\begin{equation}\label{eq:discovery_rate}
    \!\!
    DR_t = \frac{1}{N_{tests}}\sum_{j=1}^{N_{tests}} \mathds{1}_{\hat{i}_t=i^*}, \;
    \mathds{1}_{\hat{i}_t=i^*} = 
    \begin{cases}
    1, & \text{if } \hat{i}_t=i^*  \\
    0, & \text{otherwise},
    \end{cases}
\end{equation}
where $\hat{i}_t$ follows \eqref{eq:Tx_estimate}.
From a probabilistic perspective, $DR_t$ is the empirical probability of identifying Tx at time $t$ over the MC tests.

\subsubsection*{\textbf{Baselines}}
\label{subsubsec:baselines}
The following original baselines are considered, all leveraging the main PAS approach proposed here:
\begin{itemize}
    \item \textit{RaPAS}: same pipeline of Ra$^2$ViPAS, but only Rx$^{(dir)}$ is used, 
    namely $y_t = y_{\text{RF},t}$.
    \item \textit{RaViPAS}: same pipeline of Ra$^2$ViPAS, but Rx$^{(dir)}$ is not used, 
    namely $y_t = y_{D,t}$.
\end{itemize}

\subsection{Discussion}\label{subsec:discussion_numerical}

\begin{figure}[t]
\centering
\includegraphics[width=0.4\textwidth]{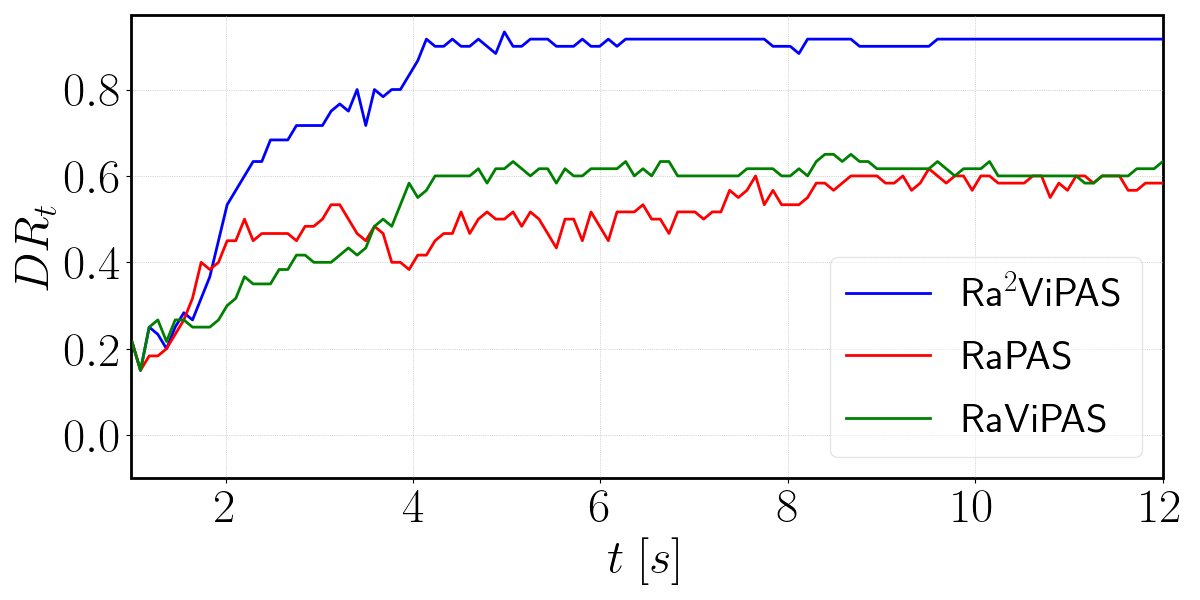}
\vspace{-0.2cm}
\caption{\footnotesize DR of Ra$^2$ViPAS, RaPAS and RaViPAS over a MC experiment.}
\label{fig:DR}
\vspace{-0.5cm}
\end{figure}


Fig. \ref{fig:DR} depicts the discovery rate of Ra$^2$ViPAS, RaViPAS and RaPAS. 
The control law \eqref{eq:control_input} induces the same qualitative behavior in all three algorithms, even though with different final results. 

Initially, the sparse sampling of the search domain $[-\pi,\pi]$ produces wrong localization solutions \eqref{eq:gamma_hat}; therefore, $DR$ is low. At the same time, the sparse domain coverage implies large uncertainties in the surrogate model; this induces an explorative platform behavior, according to \eqref{eq:ucb}. BO generates an efficient exploration process that allows the platform to focus on regions that are more likely to optimize the objective function; hence, $DR$ increases remarkably in this phase. When enough information is collected, exploitation overcomes exploration and $DR$ meets a converge. 
Fig. \ref{fig:DR} shows that Ra$^2$ViPAS has a short exploration phase ($\approx \SI{4}{\s}$) with the highest ($\approx 92\%$) and smoothest convergence behavior. This means that the information fusion process employed in Ra$^2$ViPAS produces a fast and efficient information gain, which leads to high estimation accuracy, stability and robustness.

In conclusion, the performance of Ra$^2$ViPAS justifies the higher hardware requirements w.r.t. RaPAS and RaViPAS.

\section{Conclusion}\label{sec:conclusion}
This  work  proposes  a probabilistic bi-radio-visual active sensing framework, applied to the transmitter discovery problem. We combine target visual detectability and radio signal strength into a Bayesian Optimization framework, responsible for the generation of the platform control movements. 
The approach can be extended to various application domains and in the specifically considered scenario the suggested strategy attains a $92\%$ accuracy level, $30\%$ higher than the two baselines under comparison.  
Future work will be devoted to real-life experiments and to the extension of Ra$^2$ViPAS to cluttered dynamic contexts.




%





\bibliographystyle{IEEEtran}
\bibliography{IEEEabrv,References}

\end{document}